\documentclass{article}
\usepackage{amsthm}
\usepackage{amsmath}
\usepackage{amssymb}
\usepackage{hyperref}
\usepackage{xcolor}
\usepackage{graphicx}
\usepackage{float}    
\usepackage{placeins} 

\title{
    Geometrical structures of digital fluctuations
    in parameter space of neural networks
    trained with adaptive momentum optimization
}
\author{
    Igor V. Netay
    \thanks{Joint Stock "Research and production company ``Kryptonite"}
    \thanks{
        Institute for Information Transmission Problems,
        Russian Academy of Sciences, Moscow, Russia
    }
    \href{mailto:i.netay@kryptonite.ru}{i.netay@kryptonite.ru}
}
\date{}

\begin{document}
\maketitle

\begin{abstract}
    We present results of numerical experiments for neural networks with 
    stochastic gradient-based optimization with adaptive momentum.
    This widely applied optimization has proved convergence and practical efficiency,
    but for long-run training becomes numerically unstable.
    We show that numerical artifacts are observable not only for large-scale models
    and finally lead to divergence also for case of shallow narrow networks.
    We argue this theory by experiments with more than $1600$ neural networks
    trained for $50000$ epochs.
    Local observations show presence of the same behavior of network parameters
    in both stable and unstable training segments.
    Geometrical behavior of parameters forms double twisted spirals in the parameter
    space and is caused by alternating of numerical perturbations with next 
    relaxation oscillations in values for $1^{st}$ and $2^{nd}$ momentum.
\end{abstract}

\section*{Introduction}

Stochastic gradient-based optimization with adaptive momentum was introduced 
in~\cite{KingBa15} and is currently widely applied.
It uses estimations for $1^{st}$ and $2^{nd}$ momentum with exponential
decay with parameters $\beta_1$ and $\beta_2$.
The optimization is provided by well known conditions on $\beta_1$ and $\beta_2$
to make the learning convergent.
There are also some guaranties of convergence 
(see~\cite[Theorem~4.1 and Corollary~4.2]{KingBa15}).

At the same time, it turns out that Adam optimization is unstable 
(see~\cite{fluctuations,molybog2023theory}).
The spikes of loss arise and reveal the difference between predicted 
mathematical convergence and resulting experiment results.
The main reason of the difference is that real numbers quantization introduce
inaccuracies in all arithmetic operations.
There are some approaches to minimize spikes (see~\cite{ede2020adaptive}).

Although digital noise in arithmetic operations behave irregularly, 
its influence on the network behavior gets some regular geometrical patterns.
We will show below (see~\S\ref{sec:fluct}) how do these patterns looks like.
These noisy numerical effects are the main obstacle for convergence.
The numerical instability problem is not a specificity of adaptive momentum
optimization (for instance, SGD has also some fluctuating behavior~\cite{fluctuations}).
Conversely, this observable geometrical behavior is the feature of Adam optimizer.
In~\S\ref{sec:osc} we describe unstable behavior and show that parameters can
oscillate in fragments of training with monotonic loss.

\section{Related work}
\label{sec:related}

Local training behavior of neural networks and some useful visualization
approaches is presented in~\cite{li2018visualizing} for loss functions.
Results on global neural networks landscape can be found in~\cite{Nguyen2018OnTL}.
In~\cite{JMLR:v20:18-674} behavior of single-layer networks was studied 
in terms of local landscape and spurious valleys.
A review on dynamics for different gradient descent optimizations with some
visualization can be found in~\cite{Ruder2016AnOO}.

Some theory for local spikes for adaptive momentum learning is given
in~\cite{molybog2023theory}.
There spikes are studied and found for big neural networks.
It was conjectured there that spikes appear mainly for big models for 
long-scale training.
Here we show that the same effects occur also for networks with just a few neurons.
It was also checked (see~\cite{fluctuations}) that the same happens for ``middle-sized''
well known networks like \texttt{LeNet}.
Some sufficient mathematical conditional for numerical stability of deep neural
networks can be found in~\cite{Berlyand2020StabilityFT,RePEc:hin:jnlmpe:4321312}.

We consider behavior of neural networks due to numerical errors and arising geometrical structures.
Training limitations due to numerical instability are studied 
in~\cite{fluctuations,Karner2022LimitationsON,NEURIPS2022_7b97adea,Antun2021TheDO}.
Details and problems for 16-bit network instability can be found in~\cite{Yun2023StableAO}.

\section{Structure of oscillations}
\label{sec:osc}

Analysis of the loss function landscape is an interesting part of global
neural network behavior analysis~(see~\cite{li2018visualizing}).
Here we pay attention to monotonic smooth fragments of loss.
These landscape fragments are not usually being studied in details for their
seeming simplicity.

\begin{figure}[h]
    \centering
    \includegraphics[width=6cm]{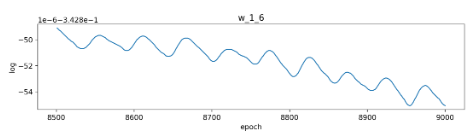}
    \includegraphics[width=6cm]{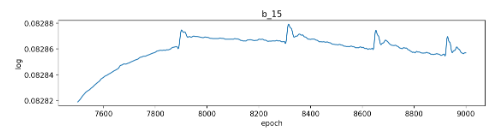}
    \includegraphics[width=6cm]{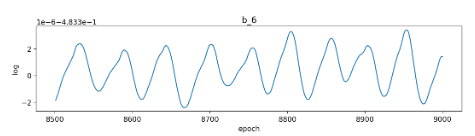}
    \includegraphics[width=6cm]{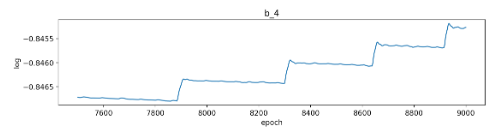}
    \includegraphics[width=6cm]{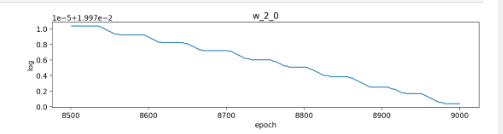}
    \includegraphics[width=6cm]{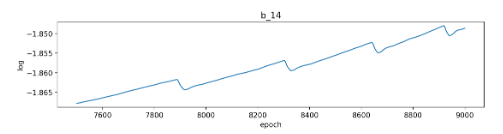}
    \includegraphics[width=6cm]{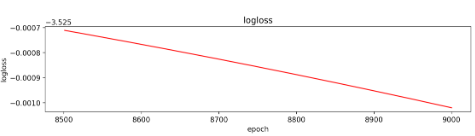}
    \includegraphics[width=6cm]{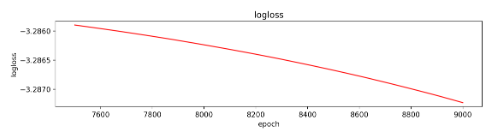}
    \caption{
        Common behavior of some pair of parameters and loss 
        for $(12, 24)$ (left) and $(16, 29)$ (right).
    }
    \label{fig:osc}
\end{figure}

Let us see how network trainable parameters behave on smooth simple 
loss fragment at a large scale (see~Fig.\,\ref{fig:osc}).

The loss value collects all the inaccuracies from the whole network parameters 
and then backpropagates them back to each trainable parameter, so the 
parameters oscillate together.
At the same time, parameters have separate exponential decaying averaging.
This causes common wave-like pseudo-periodic behavior.

Each numeric parameter should be considered as an approximation of real value
and not as a real value.
Each computational operation is a bit noisy.
So, in the neighborhood of a particular weight and bias values there is some
level of digital noise in parameters.
Exponential decay causes subtraction of close numbers, and if the difference becomes
close to the noise level, then numeric perturbations of some relatively stable level occur.
After this, the parameter values and loss exponentially tend back to convergence domain until
they get back to unstable behavior area.

This explains the small oscillations during the network training.
Actually, even for smooth monotonic approximately linear behavior, weights can
oscillate (see~Fig.\,\ref{fig:osc}).

\section{Structure of fluctuations}
\label{sec:fluct}

Actually, the fluctuating network behavior is explained likely small oscillations
with the same periods and geometric characteristics.
The only geometric difference is that the noise level grows high enough to 
imply loss explosions.

But there are meaningful corollaries:
\begin{itemize}
    \item perturbation level exceeds loss level and can make observable effects on loss values,
    \item usually, it is easy to see both ``fast'' and ``slow'' oscillations.
\end{itemize}
Here fast and slow oscillations are pseudo-periodic behavior with a big period
and periodic behavior with small period.
Their period correspondence with Adam parameters will be discussed below~(see\S\ref{sec:corr}).

\begin{figure}[h]
    \centering
    \includegraphics[width=6cm]{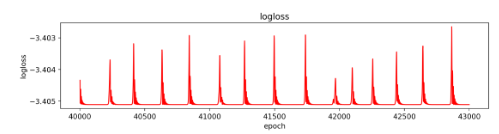}
    \includegraphics[width=6cm]{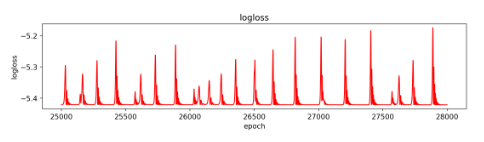}
    \includegraphics[width=6cm]{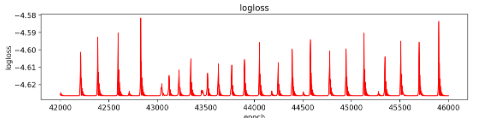}
    \includegraphics[width=6cm]{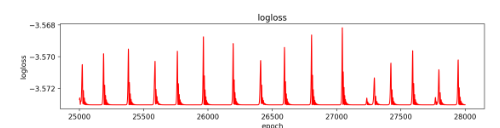}
    \caption{Loss behavior for $(17,12)$, $(20,20)$, $(26,26)$, $(27,27)$.}
    \label{fig:loss}
\end{figure}

\begin{figure}[h]
    \centering
    \includegraphics[width=8cm]{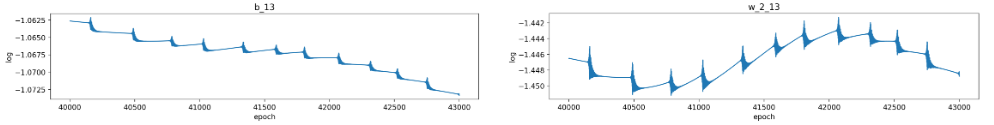}
    \includegraphics[width=4cm]{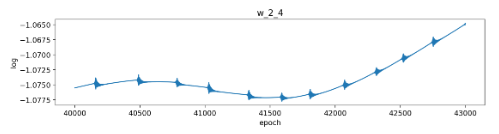}
    \includegraphics[width=4cm]{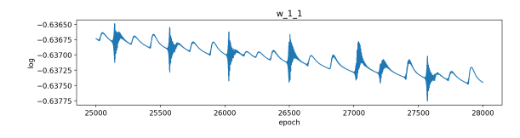}
    \includegraphics[width=8cm]{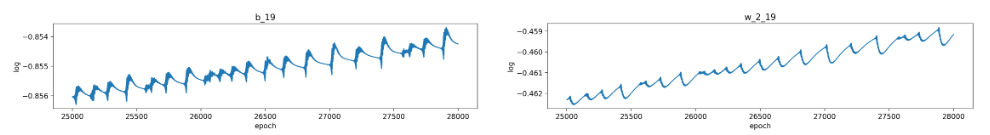}
    \includegraphics[width=12cm]{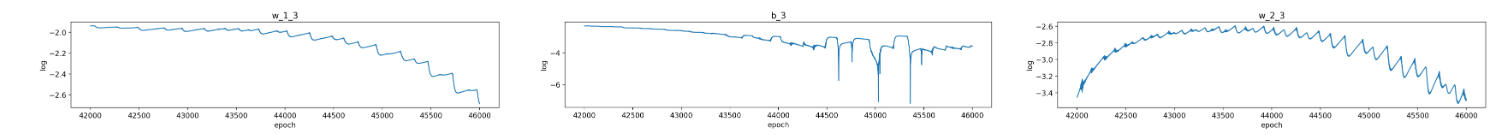}
    \includegraphics[width=12cm]{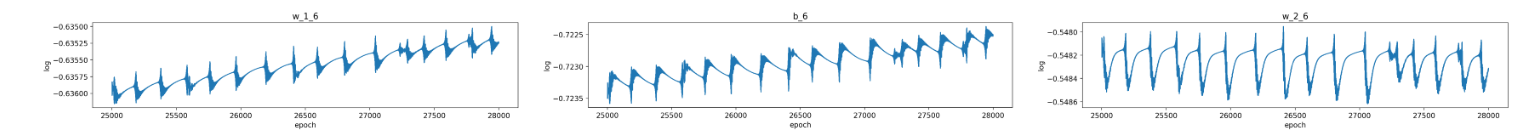}
    \caption{Behavior of some parameters for $(17,12)$, $(27,27)$, $(20,20)$, $(27,27)$.}
    \label{fig:wb}
\end{figure}

Slow oscillations of parameters occur with fluctuations of loss,
and fast oscillations can be seen in the short exponential decay of exploded loss.
See simultaneous common behavior of some weights and biases on~Fig\,\ref{fig:wb}.
(Here weights of $1^{st}$ for correspond to $1^{st}$~loss on~Fig\ref{fig:loss} and so on).

\begin{figure}[h]
    \centering
    \includegraphics[width=5.6cm]{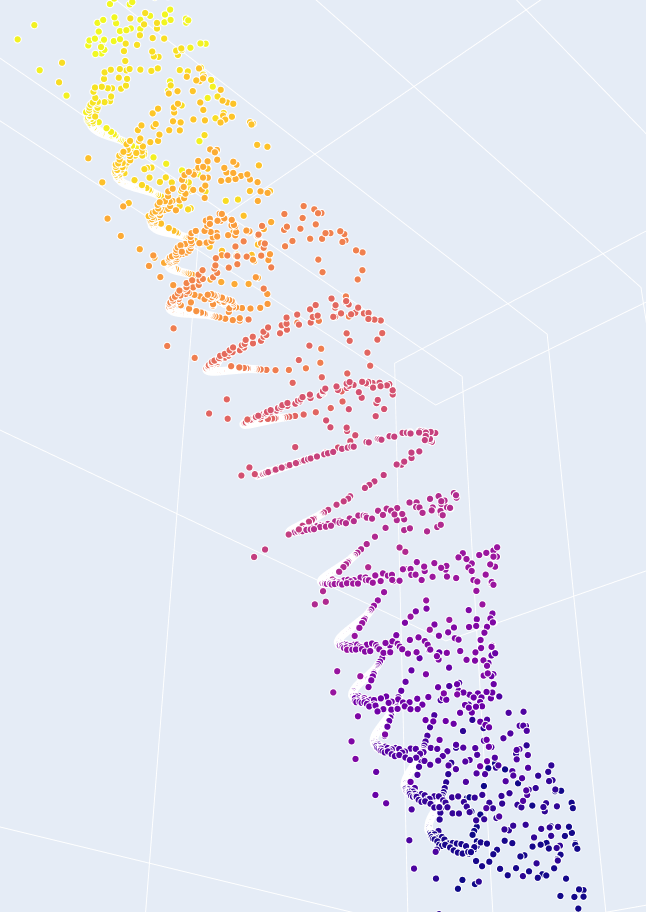}
    \includegraphics[width=3.3cm]{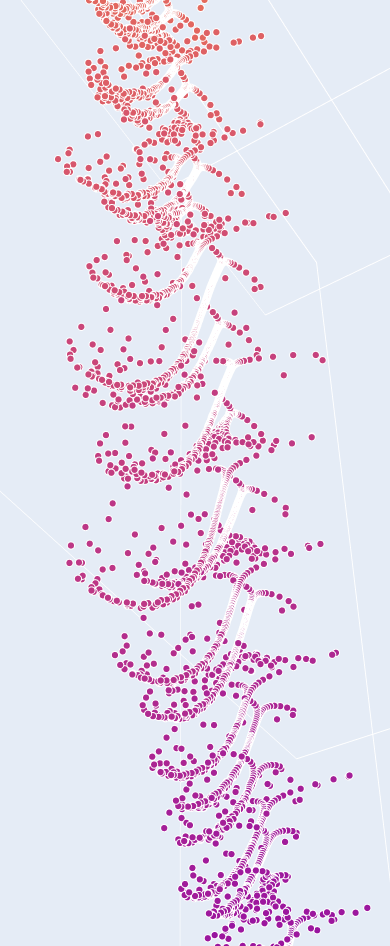}
    \caption{Common behavior of some parameters for $(26,26)$, $(27,27)$.}
    \label{fig:sp1}
\end{figure}

\begin{figure}[h]
    \centering
    \includegraphics[width=4.8cm]{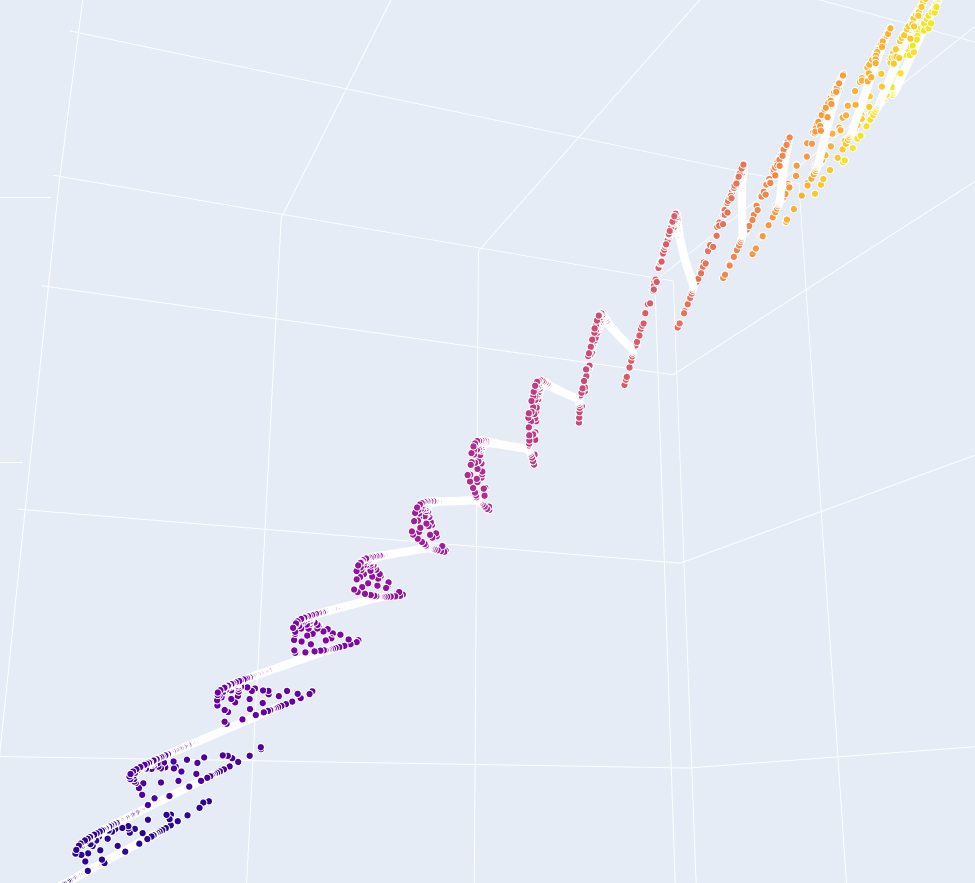}
    \includegraphics[width=7cm]{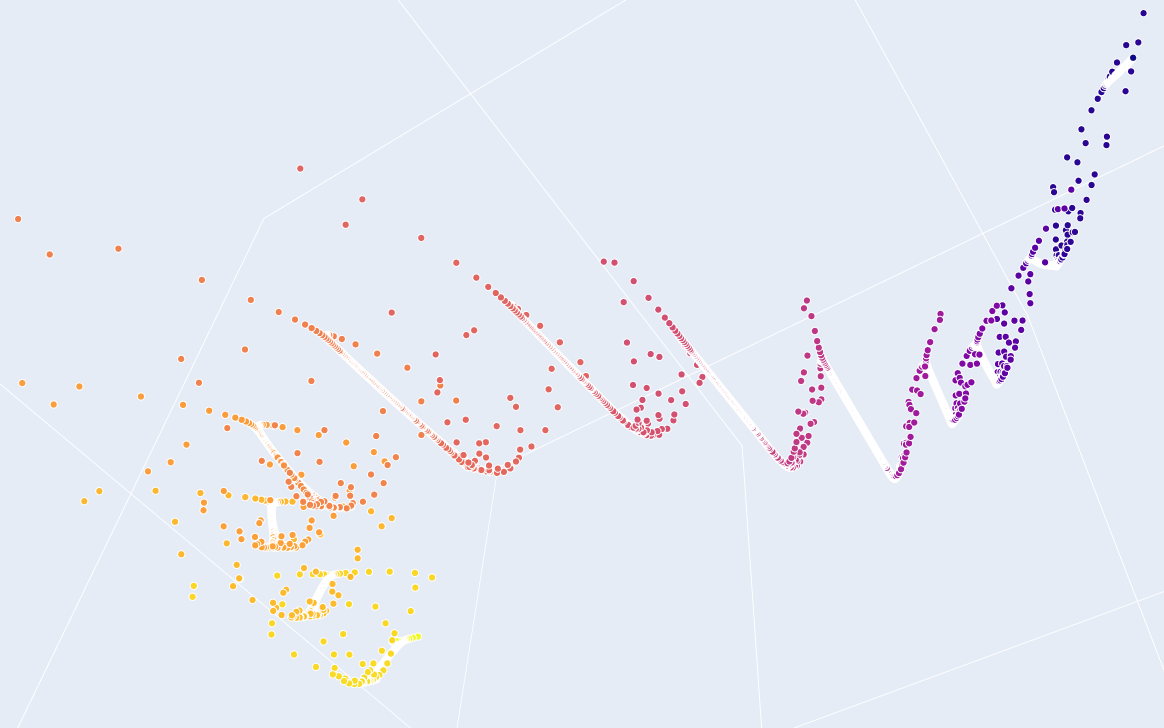}
    \includegraphics[width=12cm]{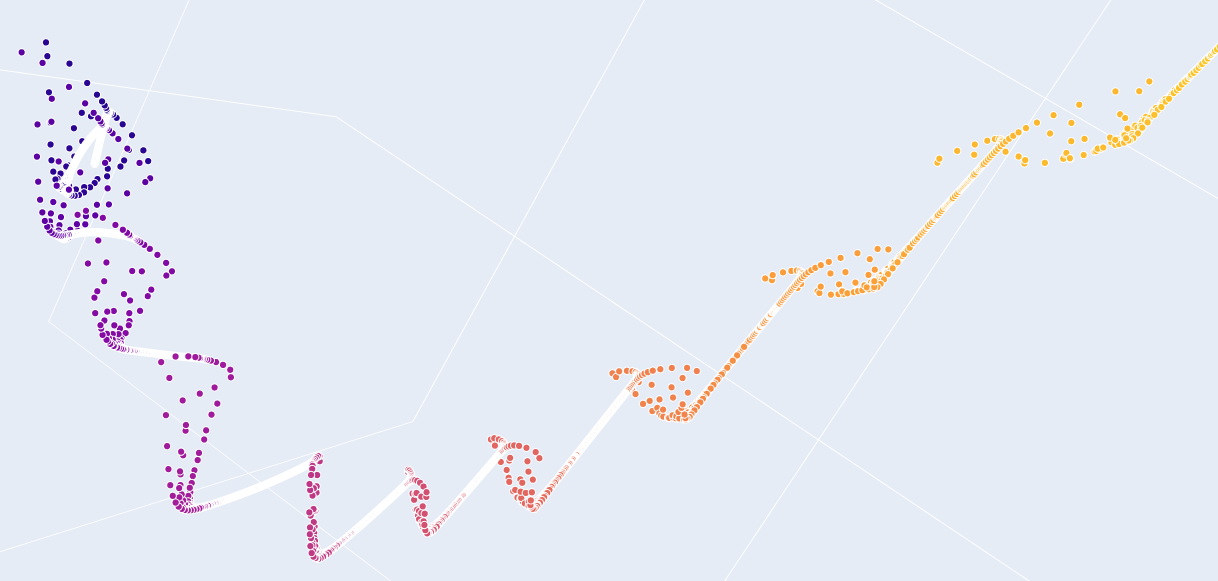}
    \caption{Common behavior of some parameters for $(21, 21)$, $(17,12)$ (2 pictures).}
    \label{fig:sp2}
\end{figure}

\begin{figure}[h]
    \centering
    \includegraphics[width=6cm]{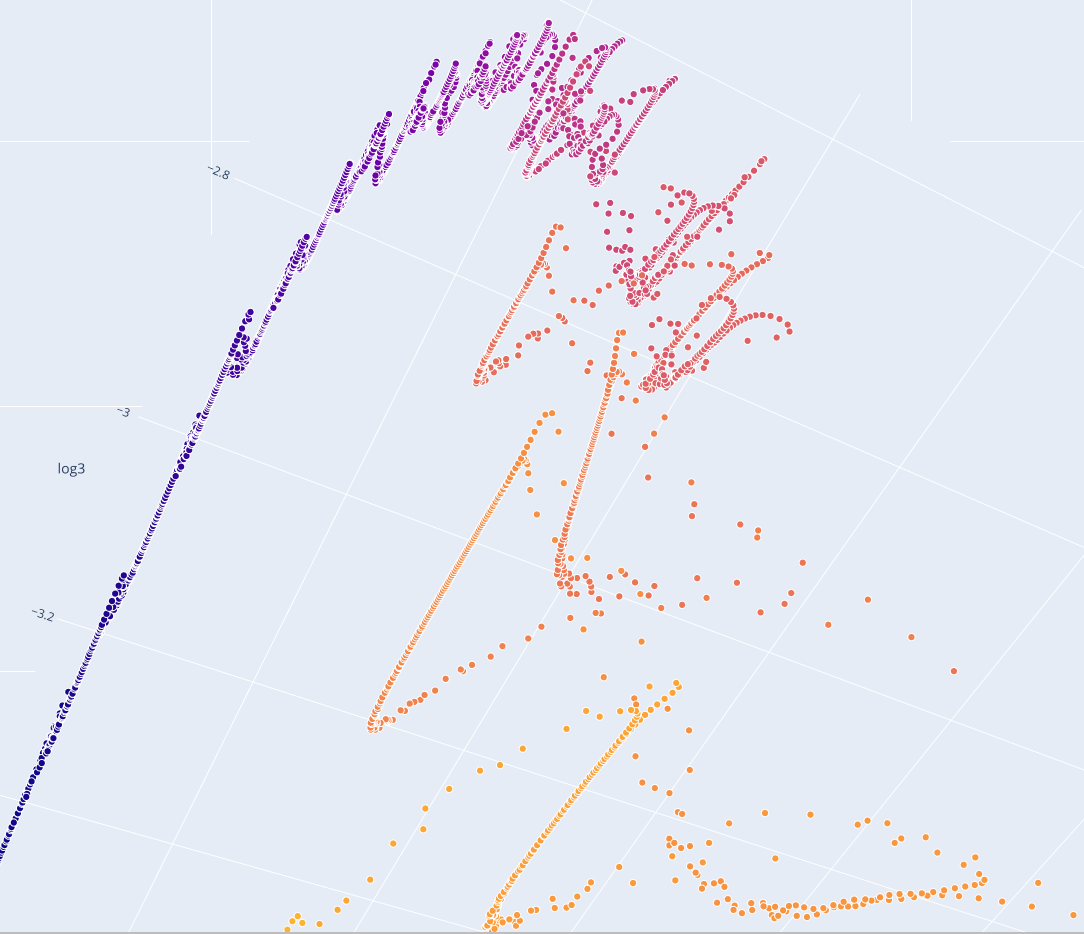}
    \includegraphics[width=6cm]{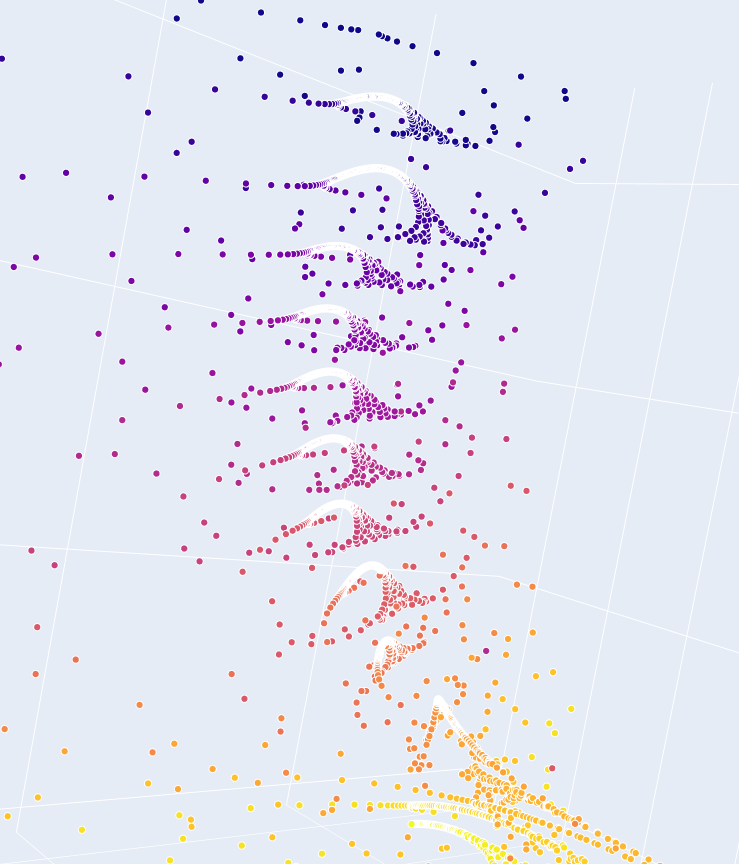}
    \caption{Common behavior of some parameters for $(20,20)$, $(34, 34)$.}
    \label{fig:sp3}
\end{figure}

We illustrate common behavior of triples of neural network parameters with
fluctuations with parametric plots (see~Fig.\,\ref{fig:sp1},\ref{fig:sp2},\ref{fig:sp3}).
These illustrations contain given above cases of parameters~(Fig.\,\ref{fig:loss})
in the same range of epochs (the number of epoch corresponds to the point color)
and some other cases for pairs of neurons number and number of dataset.

\begin{figure}[h]
    \centering
    \includegraphics[width=12cm]{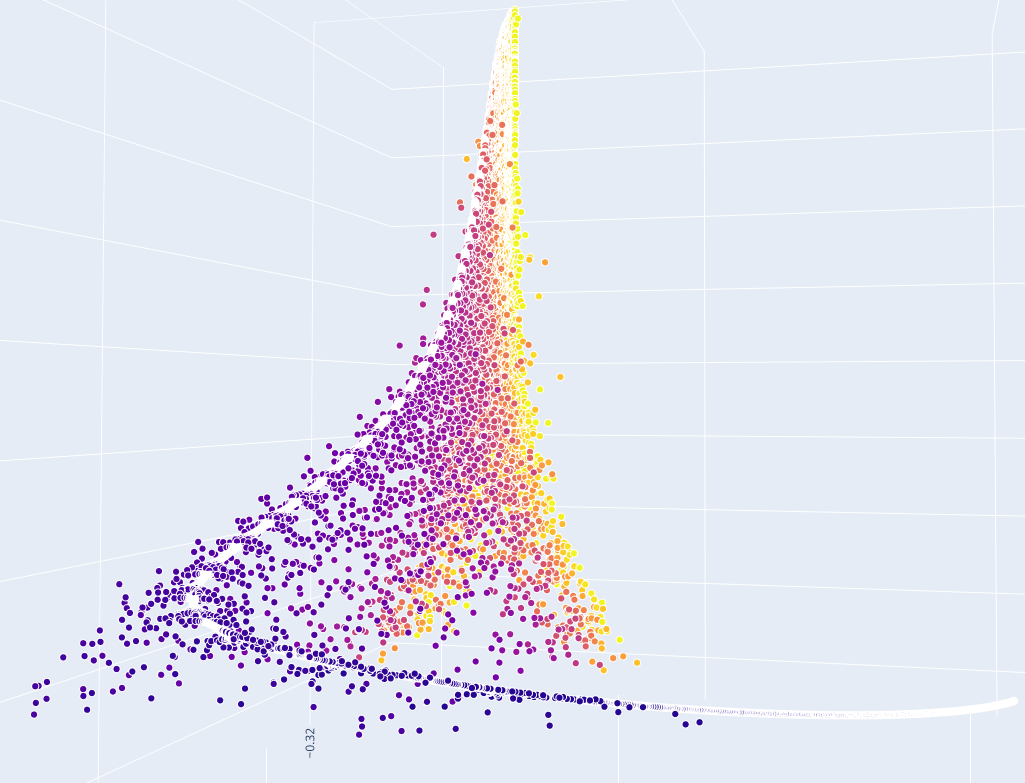}
    \caption{Common behavior of some pair of parameters and loss for $(12, 5)$.}
\end{figure}

We see that points in the parameter space form two spirals in three-dimensional 
space with opposite rotation orientation.
Slow oscillations form spirals turns.
Fast oscillations form twists of spirals in each turn, when they become close.
This corresponds to behavior inside one fluctuation.

\section{
    Correspondence between spirals parameters and optimizers hyper-parameters
}
\label{sec:corr}

It appears that parameters of spirals correspond to parameters of 
Adam optimizer~$\beta_1$ and $\beta_2$.
Fluctuations have period near to $\frac{1}{1-\beta_2}$ and fast oscillations 
inside a fluctuation have period near to $\frac{1}{1-\beta_1}$.

The notion of period here requires some more detailed explanation.
As it was mentioned above, fluctuations are rather pseudo-periodic, 
than periodic~(see examples on Fig.\,~\ref{fig:loss}).
So, the distance from a spike to the next spike is close to a linear dependence
on logarithm of spike magnitude.
The coefficient of this linear dependence changes linearly with~$\frac{1}{1-\beta_2}$.

\section{Conclusion}
\label{sec:conclusion}

For experiments $1600$ neural networks were trained for $50000$ epochs each one.
We can see common geometrical patterns for groups of network parameters.

For training neural network with adaptive momentum optimizers, possible periods of fluctuations
can be estimated, and fluctuations can be predicted before they achieve high amplitude.
Local trainable parameter behavior analysis could be a useful tool for
numerical stability estimation.

Although there are some estimations and proof of convergence for some cases
(see~\cite[Corollary~4.2]{KingBa15}), we see in experiments that even simple
one-layer networks fail to converge.
This means that mathematical model fails, and computations are just inexact.
So, the main reason of global divergence lies in numerical inexactness.

Actually, there is a way to automagically estimate numerical 
inaccuracies~\cite{netay2024algorithms} for particular tensor elements during 
computations with floating point numbers.

\section{Future work}
\label{sec:future}

It would be interesting to study dynamics of training and fluctuations for other
popular optimizers like SGD.

\section{Acknowledgements}

The author is grateful to his Kryptonite colleagues Vasily Dolmatov,
Dr. Nikita Gabdullin and Dr. Anton Raskovalov for fruitful discussions of topic and results.

\FloatBarrier

\bibliographystyle{unsrt}
\bibliography{refs}

\end{document}